\documentclass[twoside]{article}

\usepackage[accepted]{aistats2026}
\usepackage{xcolor}
\usepackage{booktabs,tabularx,adjustbox}
\usepackage{amssymb}
\usepackage{tabularx}
\usepackage{dblfloatfix}
\usepackage{graphicx}
\usepackage{lipsum}
\usepackage{siunitx}
\newcolumntype{Y}{>{\raggedright\arraybackslash}X} 

\usepackage[round, authoryear]{natbib}

\begin{document}

%

%

\twocolumn[

\aistatstitle{FedGTEA: Federated Class-Incremental Learning with Gaussian Task Embedding and Alignment}

\aistatsauthor{ Haolin Li \And Hoda Bidkhori}

\aistatsaddress{ 
George Mason University  \\ 
\texttt{hli54@gmu.edu} 
\And
George Mason University \\
\texttt{hbidkhor@gmu.ed}u } ]

\begin{abstract}

    We introduce a novel framework for Federated Class Incremental Learning, called Federated Gaussian Task Embedding and Alignment (FedGTEA). FedGTEA is designed to capture task-specific knowledge and model uncertainty in a scalable and communication-efficient manner. At the client side, the Cardinality-Agnostic Task Encoder (CATE) produces Gaussian-distributed task embeddings that encode task knowledge, address statistical heterogeneity, and quantify data uncertainty. Importantly, CATE maintains a fixed parameter size regardless of the number of tasks, which ensures scalability across long task sequences. On the server side, FedGTEA utilizes the 2-Wasserstein distance to measure inter-task gaps between Gaussian embeddings. We formulate the Wasserstein loss to enforce inter-task separation. This probabilistic formulation not only enhances representation learning but also preserves task-level privacy by avoiding the direct transmission of latent embeddings, aligning with the privacy constraints in federated learning. Extensive empirical evaluations on popular datasets demonstrate that FedGTEA achieves superior classification performance and significantly mitigates forgetting, consistently outperforming strong existing baselines.
 
\end{abstract}

\section{Introduction}

In this paper, we propose a new algorithm for Federated Class-Incremental Learning (FCIL) \citep{birashkFederatedContinualLearning2025} that effectively models task-level knowledge to enable scalable and privacy-preserving model aggregation. FCIL is an emerging field of study that addresses both the problem of learning from statistically heterogeneous clients \citep{wuerkaixiAccurateForgettingHeterogeneous2025} and memorizing previously learned tasks \citep{qiBetterGenerativeReplay2023} without catastrophic forgetting. This paradigm offers a powerful solution that inherently preserves user privacy at both data and task levels \citep{birashkFederatedContinualLearning2025}.

As a hybrid of Federated Learning (FL) and Class Incremental Learning (CIL), FCIL has received increasing attention due to its ability to learn continuously in a distributed manner \citep{birashkFederatedContinualLearning2025}. Leveraging the decentralized nature of FL, FCIL systems can benefit from large volumes of data generated by clients \citep{dongFederatedClassIncrementalLearning2022a}, while also accommodating a dynamically expanding label space without catastrophic forgetting \citep{qiBetterGenerativeReplay2023}. These characteristics make FCIL a practical and promising framework for real-world machine learning applications. FCIL introduces three core challenges \citep{birashkFederatedContinualLearning2025}: 1) Catastrophic Forgetting - Occurring both locally on clients and during global aggregation. 2) Statistical Heterogeneity - Data distributions are typically non-IID across clients. 3) Task Context Ambiguity - The absence of task identity at test time leads to semantic drift and performance degradation. 

Many works in FCIL aim at advancing memory-based methods. They mostly focus on exploiting data-level features \citep{birashkFederatedContinualLearning2025}. For example, the GLFC algorithm \citep{dongFederatedClassIncrementalLearning2022a} uses an exemplar memory at clients with a proxy server. On the other hand, the FLwF-2T approach \citep{usmanovaDistillationbasedApproachIntegrating2021} focuses on knowledge distillation to transfer knowledge between models. Meanwhile, the FedCIL method \citep{qiBetterGenerativeReplay2023} incorporates generative replay with an additional model consolidation step. In contrast, our proposed algorithm FedGTEA extends beyond data-level knowledge by also modeling task-level context. Specifically, FedGTEA enables client models to generate Gaussian task embeddings that will be utilized by the server. The server then leverages the Wasserstein distance to perform effective regularization, which both reduces catastrophic forgetting and promotes inter-task separation.

\begin{figure}[t]
    \centering
    \includegraphics[width=\linewidth]{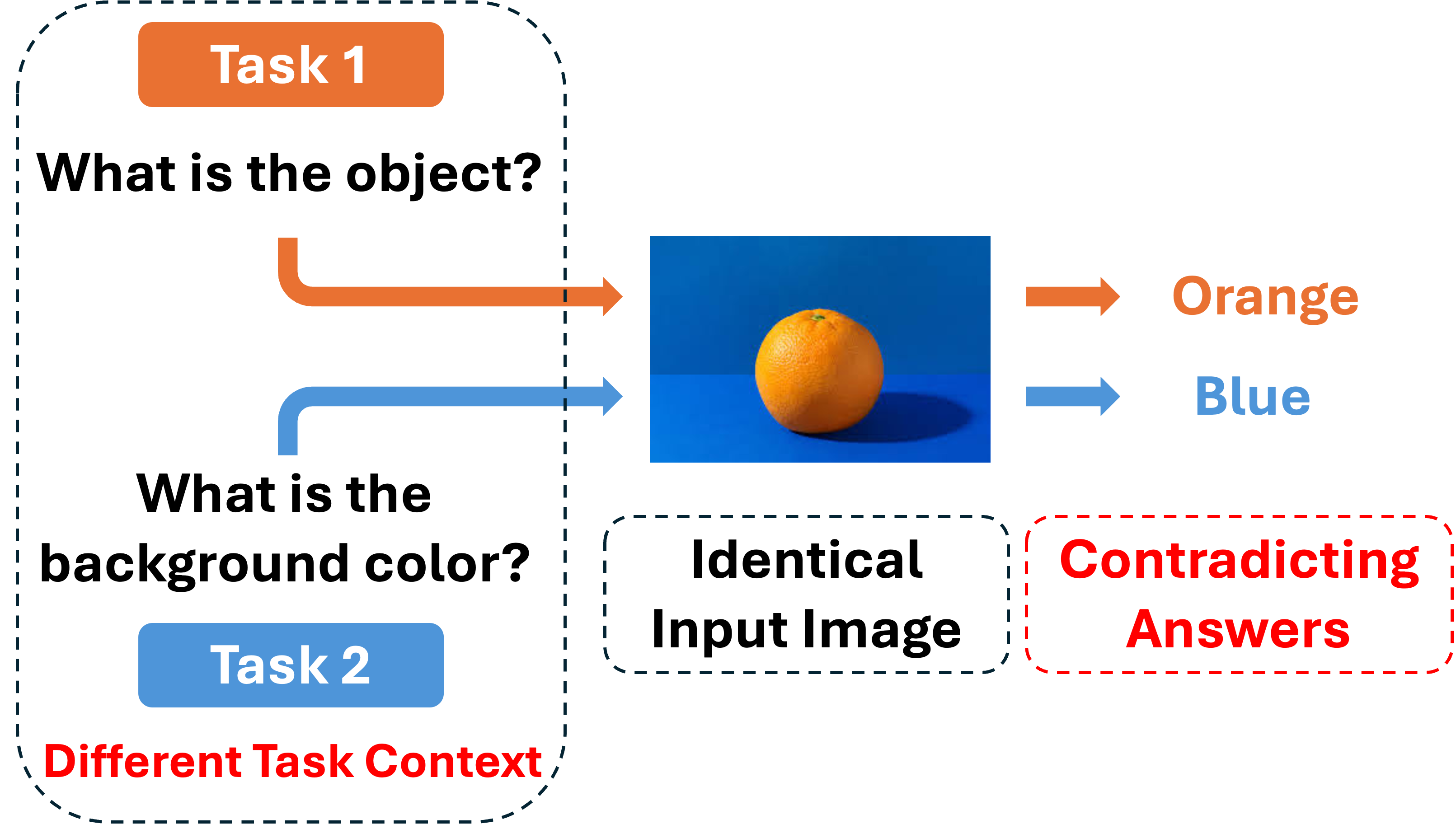}
    \caption{Given different task contexts, identical input can yield contradicting answers. Multi-task models need task-level context to process different tasks accurately.  }
    \label{fig:intro_fig}
\end{figure}

Numerous studies \citep{caruana1997multitask, achilleTask2VecTaskEmbedding2019, zamirTaskonomyDisentanglingTask2018} have argued that task-level signals provide critical context and explain how identical inputs may yield different or even contradicting outputs across different tasks. As an example in Figure \ref{fig:intro_fig}, the interpretation of the same image under different tasks (e.g. What is the object? v.s. What is the background color?) requires different task-level contextual information \citep{zamirTaskonomyDisentanglingTask2018} beyond the data-level knowledge from input itself.

To the best of our knowledge, the majority of research in this area focuses on efficient exploitation of data-level information. Only a few works in prompt learning \citep{luoFederatedClassincrementalLearning2025, bagweFedCPromptContrastivePrompt2023} incorporated task knowledge. As a result, the literature on how to effectively leverage task context in a scalable manner in FCIL is still relatively underdeveloped. Therefore, in this work, we focus on developing a memory-efficient task encoder that effectively extracts task-level context with a fixed number of parameters. On top of that, additional regularization and means of information transmission also need attention for a robust and privacy-aware FCIL system.

To extract and leverage task-level knowledge in an efficient, robust, and private manner, we propose Federated Gaussian Task Embedding and Alignment (FedGTEA). FedGTEA is an FCIL framework that integrates task context in a parameter-efficient design. We introduce a task embedding module called Cardinality-Agnostic Task Encoder (CATE), which is capable of inferring a compact task embedding from a batch of data, irrespective of the batch size. 

A key distinction of our approach is modeling the task embeddings produced by CATE as Gaussian random variables. This enables principled reasoning about uncertainty, distributional variability, and alignment. On top of this, the server performs standard federated aggregation and a task alignment and model consolidation step, utilizing the 2-Wasserstein distance to regularize task representations both spatially (across clients) and temporally (across tasks). These mechanisms collectively enhance inter-task separation and support robust model consolidation.

\noindent \textbf{Contributions.}

\begin{itemize}
    \item We propose FedGTEA, an algorithm that effectively captures task-level knowledge in a scalable and robust manner for FCIL. We introduce the Cardinality-Agnostic Task Encoder (CATE) in the client model to produce task embeddings, model these embeddings as Gaussian random variables, and leverage the 2-Wasserstein distance on the server to promote inter-task separation.

    \item The CATE module in the client model infers task embeddings from a batch of data, regardless of its size. This makes it cardinality-agnostic. By modeling the embeddings as Gaussian random variables, we enable the server to quantify inter-task distances using the 2-Wasserstein metric.

    \item On the server side, we first perform initial model aggregation using FedAvg \citep{mcmahanCommunicationEfficientLearningDeep2017} principles. Then, we formulate an optimization problem with three loss components: (i) knowledge distillation loss to transfer prior knowledge to the new global model, (ii) Wasserstein loss to promote inter-task separation, and (iii) anchor loss to limit excessive drift from the initial aggregated model. We solve this optimization via gradient descent to obtain the final global model.
    
    \item Compared to popular baselines such as (AC-GAN + ) FedAvg, (AC-GAN + ) FedProx,  GLFC, FedCIL, and FLwF-2T, FedGTEA achieves superior performance in terms of both accuracy and forgetting, with consistently low variance across all three task settings. These results highlight the effectiveness of our CATE design and Wasserstein-based regularization.

\end{itemize}

\section{Background}

In this section, we first review related works on the two core components of FCIL: Class Incremental Learning and Federated Learning, together with their assumptions. Then, we organize FCIL methods into three representative categories and discuss the literature within each. In the end, we highlight why our proposed algorithm, FedGTEA, addresses current research gaps by incorporating task-level context in a robust and privacy-aware manner.  

\noindent \textbf{Class Incremental Learning.} Class incremental Learning is one of the key paradigms in continual learning\citep{masanaClassIncrementalLearningSurvey2023a}. CIL models learn an expanding label space while assuming prohibited access to data points of previous tasks. Recent advances in CIL broadly fall into three families:
(I) \emph{Replay-based methods} like iCaRL\citep{rebuffiICaRLIncrementalClassifier2017} and GEM \citep{Lopez-pazGradientEpisodicMemory2017} store an exemplar buffer and periodically rehearse it. DER \citep{buzzegaDarkExperienceGeneral2020} further leverages logits or intermediate activations to retain “dark knowledge”.
(II) \emph{Regularization-based approaches} \citep{liLearningForgetting2018} attempt to constrain model parameter updates by distilling knowledge from previous model states. 
(III) \emph{Prompt-based methods}, including L2P \citep{wangLearningPromptContinual2022} and DualPrompt \citep{ wangDualPromptComplementaryPrompting2022}, learn a pool of context vectors (prompts) to condition the model on previous tasks. Prompts are typically formed at training and selected at inference \citep{smithCODAPromptCOntinualDecomposed2023}.  

\noindent \textbf{Federated Learning.} 
Federated Learning enables decentralized training across distributed clients under strict data privacy constraints. Client data should never be shared with the server or other clients \citep{kairouzAdvancesOpenProblems2021}. To address the statistical heterogeneity across clients, popular aggregation choices are FedAvg \citep{mcmahanCommunicationEfficientLearningDeep2017} and FedProx \citep{liFederatedOptimizationHeterogeneous2020}. FedAvg averages client models based on the number of data points in client datasets, while FedProx regularizes clients' local updates with the previous global model. To secure user privacy, \citet{geyerDifferentiallyPrivateFederated2018} discussed differential privacy strategies and \citet{bonawitzPracticalSecureAggregation2017} proposes a practical cryptographic protocol.

\noindent \textbf{Federated Class Incremental Learning.} 
FCIL combines the challenges of both FL and CIL. It aims at training a global model where participating clients observe heterogeneous datasets of different tasks \citep{birashkFederatedContinualLearning2025}. 
Contemporary FCIL methods can be classified into several categories:
(I) \emph{Replay-based methods} use local exemplar buffers to retain knowledge from previous tasks \citep{liEfficientReplayFederated2024, zhangTARGETFederatedClassContinual2023a}. Some works \citep{qiBetterGenerativeReplay2023} also uses GAN modules like AC-GAN \citep{odenaConditionalImageSynthesis2017} for generative replay. 
(II) \emph{Regularization} \& \emph{distillation} approaches transfer older knowledge to the current state via server-assisted or peer-to-peer knowledge distillation \citep{zhangTARGETFederatedClassContinual2023a, babakniyaDataFreeApproachMitigate2023}. This reduces forgetting without storing data in an external memory.
(III) \emph{Prompt-based FCIL} introduces prompt pools stored on the client side to encode task context \citep{bagweFedCPromptContrastivePrompt2023, luoFederatedClassincrementalLearning2025}. Prompt-based methods effectively capture the features of clients, with the trade-off of increased memory usage and computational overhead.


Complementing current works in FCIL, our proposed FedGTEA uses the task embedding module CATE to infer task-level knowledge. It operates with a fixed number of parameters regardless of the number of tasks. This significantly alleviates memory and computation overheads. In addition to this, our Gaussian construction on the task embeddings enables us to quantify the gap between tasks using the Wasserstein distance \citep{peyreComputationalOptimalTransport2020}. Again, with Wasserstein distance, we formulate the task consolidation step to regularize the aggregated global model to both promote inter-task separation and review knowledge from previous tasks. Overall, the system shares task-level information without transmitting raw embeddings, which complies with privacy concerns in federated systems \citep{kairouzAdvancesOpenProblems2021}.

\begin{figure*}[t] 
  \centering
  \includegraphics[width=0.85\textwidth]{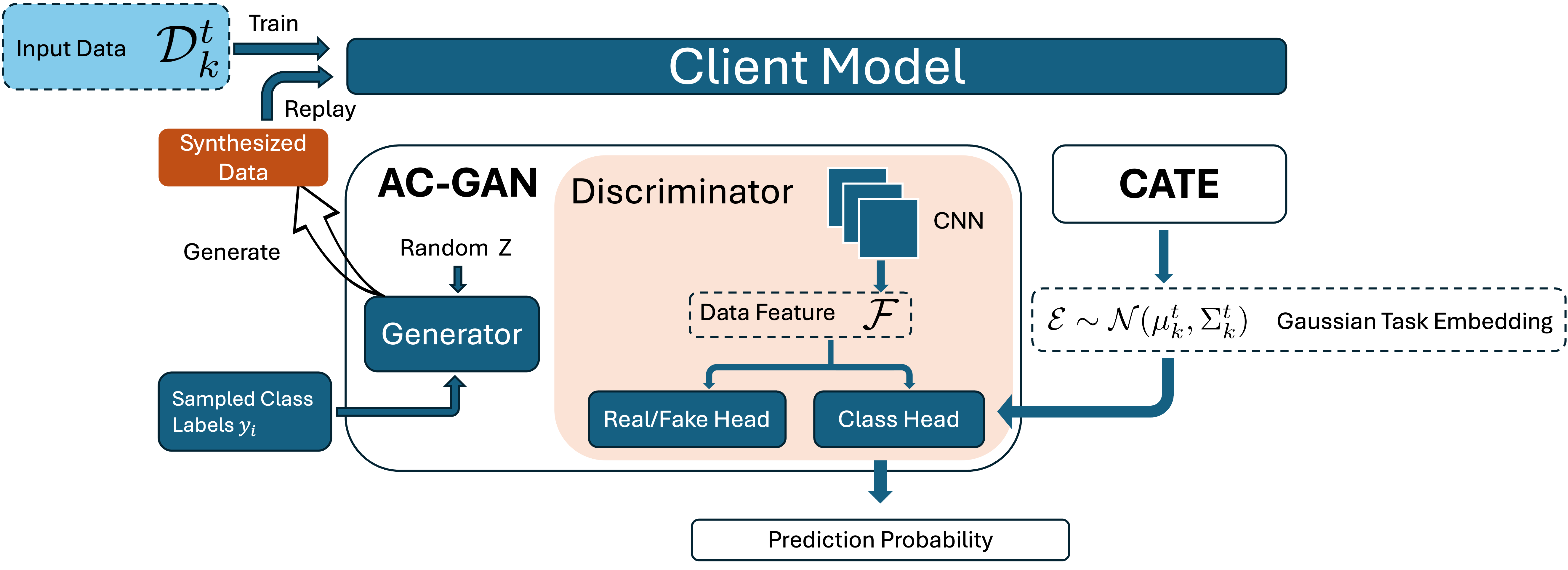} 
  \caption{This illustrates how client $\mathcal{C}_k$ is trained over task $\mathcal{T}^t$. Given locally collected dataset $\mathcal{D}_k^t$, CATE and CNN in the discriminator extract task embedding $\mathcal{E}$ and data feature $\mathcal{F}$ respectively. The class head outputs the prediction probability using both $\mathcal{F}$ and $\mathcal{E}$. At the same time, the Real/Fake head in the discriminator classifies whether the input image is genuine or synthesized. After each local iteration, $G$ generates fake images for replay purposes.  }
  \label{fig:client_model}
\end{figure*}
\section{Problem Formulation}

This section introduces the notations and the formulations of research topics in CIL, FL, and FCIL. This includes both the general descriptions of these research areas and their mathematical objectives. 

\noindent \textbf{Class Incremental Learning.}
An CIL model learns a sequence of tasks $\mathcal{T} = \{ \mathcal{T}^1, \mathcal{T}^2, \ldots, \mathcal{T}^T \}$, where $T$ is the number of tasks, and $\mathcal{T}^t$ refers to the $t$-th task. Each task $\mathcal{T}^t$ has dataset $\mathcal{D}^t$ with $n^t$ data points. The critical feature of CIL is that the label space is always increasing. 
At task $\mathcal{T}^t$, the model only have access to the current dataset $\mathcal{D}^t$, but not from previous datasets $\mathcal{D}^{t'}$, $1 \leq t' < t$. Let $\theta^t$ be the model's parameter at task $\mathcal{T}^t$, then the objective of CIL is to effectively both learn the current task and preserve performance over all previous tasks:
\begin{equation}
    \min_{\theta^t} [ \mathcal{L}(\theta^t, \mathcal{D}^1), \mathcal{L}(\theta^t, \mathcal{D}^2), \cdots, \mathcal{L}(\theta^t, \mathcal{D}^t) ]
\end{equation}
where $\mathcal{L}(\theta, \mathcal{D})$ is the loss function evaluating the empirical risk of model $\theta$ over dataset $\mathcal{D}$. 

\noindent \textbf{Federated Learning.}
A standard Federated Learning system consists of a set of $N$ different clients $\mathcal{C} = \{ \mathcal{C}_1, \mathcal{C}_2, \ldots, \mathcal{C}_N \}$ and a central aggregator (server). The FL system only handles one single task $\mathcal{T}$, but each client $\mathcal{C}_k$ collects its local dataset $\mathcal{D}_k$ and trains its model $\theta_k$ locally. At each communication round, the server aggregates a global model $\theta_g$ from the selected pool of client models. The objective of client $\mathcal{C}_k$ is:
\begin{equation}
    \min_{\theta_k} \mathcal{L}(\theta_k, \mathcal{D}_k)
\end{equation}
The objective of the FL system is to find the global model $\theta_g$ that minimizes the overall loss:
\begin{equation}
    \min_{\theta_g} [ \mathcal{L}(\theta_g, \mathcal{D}_1), \mathcal{L}(\theta_g, \mathcal{D}_2), \ldots, \mathcal{L}(\theta_g, \mathcal{D}_N) ]
\end{equation}

\noindent \textbf{Federated Class Incremental Learning.} Federated Class Incremental Learning is a special case of FL framework whose clients are CIL clients with a global task sequence $\mathcal{T}$. For each client, dataset $\mathcal{D}_k^t$ is collected by client $\mathcal{C}_k$ for task $\mathcal{T}^t_k$, which is a subset of the global task $\mathcal{T}^t$ ($\mathcal{T}_k^t \subset \mathcal{T}^t$). 
The objective of FCIL is to find the global parameters $\theta_g^t$ at task $\mathcal{T}^t$ that minimizes the loss over all seen tasks and all clients:
\begin{equation}
    \min_{\theta_g^t} [ \mathcal{S}_1, \mathcal{S}_2, \ldots, \mathcal{S}_N ]
\end{equation}
where $\mathcal{S}_k = [ \mathcal{L}(\theta_g^t, \mathcal{D}_k^1), \mathcal{L}(\theta_g^t, \mathcal{D}_k^2), \ldots, \mathcal{L}(\theta_g^t, \mathcal{D}_k^T) ]$.

\section{Methodology}

\begin{figure*}[t] 
  \centering
  \includegraphics[width=0.95\textwidth]{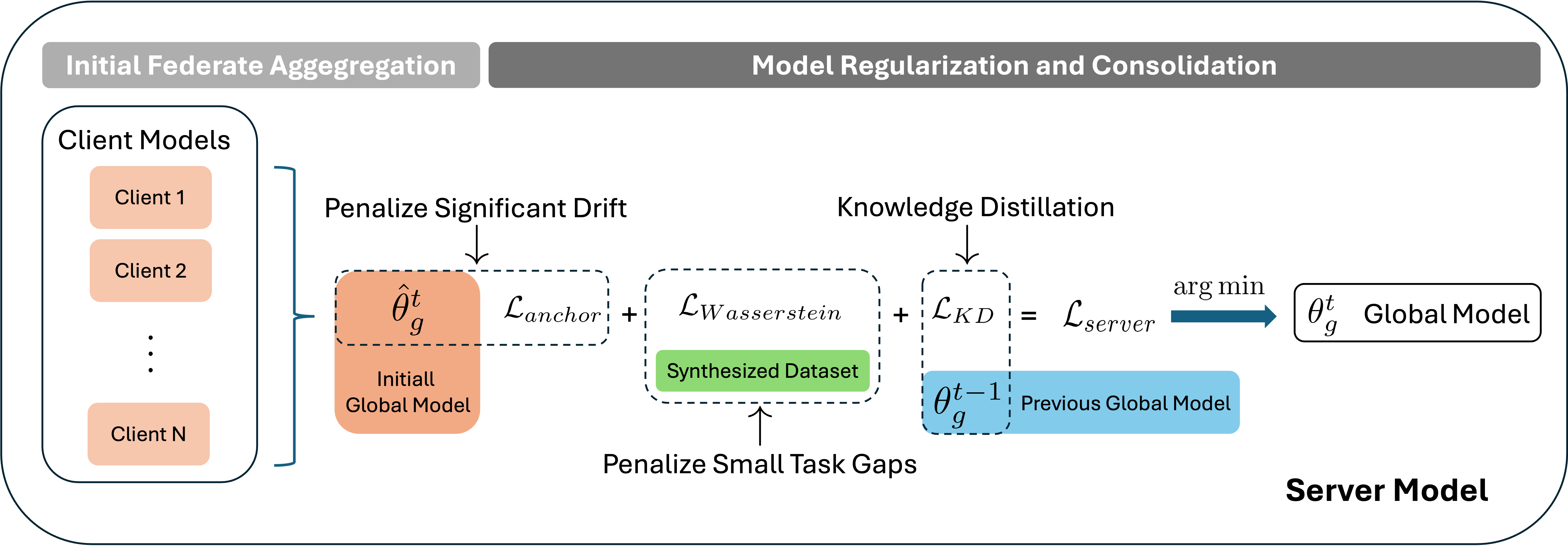} 
  \caption{This figure illustrates the aggregation and model regularization step on the server. We first obtain an initial aggregated model $\hat{\theta}_g^t$ by typical federate aggregation. Then, the regularization and consolidation use the server loss that comprises three components. Anchor loss prevents the global model from drifting too far away; Wasserstein loss promotes inter-task separation; knowledge distillation loss mitigates catastrophic forgetting.}
  \label{fig:client_model}
\end{figure*}

In this section, we first give an overview of FedGTEA. Then, we explain the core components of the client model and regularization on the server. 

\subsection{FedGTEA Overview}

\noindent \textbf{Client Model. }As shown in Figure 2, our client model has two essential components: AC-GAN, which encodes data features and replays, and CATE, which infers task knowledge. CATE infers task context from a batch of data points and outputs a Gaussian distributed task embedding $\mathcal{E} \sim \mathcal{N}(\mu, \Sigma)$. We constructed $\mathcal{E}$ as Gaussian random variables because 1) it helps us model the randomness in the task knowledge inference using different batches of data; 2) 2-Wasserstein distance can be used to later measure the gap between different tasks in a closed form; 3) it enables us to control the position and variation pattern of task embeddings via regularization on the server. At task $\mathcal{T}^t$, client $\mathcal{C}_k$ trains its own CATE locally, capturing the personalized task knowledge. In addition to the task-level information, AC-GAN provides data features $\mathcal{F}$ to assist classification. The class head fuses both the data- and task-level knowledge to decide which class a given image belongs to. Besides knowledge fusion, the discriminator's Real or Fake head distinguishes whether a given image is genuine or synthesized, enabling the generator $G$ to learn the underlying data feature distribution. By the end of each training step, the generator $G$ synthesizes images with classes sampled from the current label space for replay purposes. We note that all clients share the same CATE architecture. 

\noindent \textbf{Server Aggregation \& Regularization. }After client $\mathcal{C}_k\; (k=1,\ldots,N)$ finishes training on task $\mathcal{T}^t$ locally, its model parameters $\theta_k^t$ are then uploaded to the server for a new round of aggregation. Following the principle of FedAvg \citep{mcmahanCommunicationEfficientLearningDeep2017}, the server first obtains the initial global model 
\begin{equation}
    \hat{\theta}_g^t = \sum_{k=1}^N w_k\theta_k^t
\end{equation}
with weights $w_k$ being proportional to the number of local data points $|D_k^t|$. Next, to mitigate forgetting and promote robust task separation, we formulate a model consolidation step that optimizes $\hat{\theta}_g^t$ based on the server loss $\mathcal{L}_{server}$. The server then obtains the final global model 
\begin{equation}
    \theta_g^t = \arg \min_{\theta \in \Theta} \mathcal{L}_{server}
\end{equation}
and distributes $\theta_g^t$ to all clients as the starting point of the next round of training. The server loss is the weighted sum of three loss functions:
\begin{equation}
    \mathcal{L}_{server} = \alpha \mathcal{L}_{KD} + \beta \mathcal{L}_{Wasserstein} + \gamma \mathcal{L}_{anchor}
\end{equation}
Here we use knowledge distillation loss $\mathcal{L}_{KD}$ to transfer older knowledge from the previous global model to the current global model. Wasserstein loss $\mathcal{L}_{Wasserstein}$ penalizes small gaps between tasks, thus promoting inter-task separations. Furthermore, the anchor loss $\mathcal{L}_{anchor}$ uses $L_2$ norm to prevent significant drift from the anchor model $\hat{\theta}_g^t$.

\subsection{FedGTEA: Client Model} 
The client model consists of a CATE module with AC-GAN. First, we introduce our CATE design and Gaussian construction that handles task knowledge extraction. Then, we discuss briefly about AC-GAN. In the end, we explain the client training pipeline. 

\noindent \textbf{Cardinality-Agnostic Task Encoder. }
A task encoder generates meaningful task embeddings that the model can use to condition itself on. It improves model performance by providing task-level context. 

In our implementation, we design CATE as a fully connected neural network. Given a batch of data $B$, it outputs one task embedding in $\mathbb{R}^d$. Specifically, given any input batch $B =(x_1, x_2, \ldots, x_b)$ with $b=|B|$ as batch size, the task embedding $\mathcal{E}_B$ is computed as:
\begin{equation}
    \mathcal{E}_B = \frac{1}{b} \sum_{i=1}^b CATE(x_i) = \frac{1}{b} \sum_{i=1}^b \mathcal{E}_i \in \mathbb{R}^d
\end{equation}

As a function, CATE infers task knowledge using any number of input data points. In fact, with more data points, CATE should yield more accurate task embeddings. This is the reason why it's Cardinality-Agnostic. In addition, the number of parameters in CATE does not grow with the number of tasks, which makes it scalable for long task sequences. 

Mathematically, we define CATE as a function $f(\cdot)$ that is capable of mapping a batch of any size to $\mathbb{R}^d$, where $d$ is the dimension of the task embedding space:
\begin{align*}
    f: \bigcup_{n \geq 1} \underbrace{D \times D \times \cdots \times D}_{n\; times} \rightarrow &\mathbb{R}^d \\
    \bar{x} = (x_1, x_2, \ldots, x_n) \mapsto  f(\bar{x})=&E
\end{align*}
here $\times$ is the Cartesian product.

It is important to remark that CATE can be designed in various ways and can be more complicated, including CNN-based architectures. Yet in this paper, we show that even a simple design yields meaningful improvements. 

\noindent \textbf{Gaussian Task Embedding. }Task embeddings are the vector outputs of CATE. We construct the task embedding inferred by CATE as a Gaussian random variable. Globally, given the distribution $\mathcal{D}^t$ of task $\mathcal{T}^t$, we model $\mathcal{E}^t = CATE(x),\; x \sim D^t$ as a Gaussian random variable
\begin{equation*}
    \mathcal{E}^t \sim \mathcal{N}(\mu^t, \Sigma^t)
\end{equation*}
with position $\mu^t$ and variation pattern $\Sigma^t$. For each client $\mathcal{C}_k$, given different local data distribution $\mathcal{D}_k^t$, client-specific task embedding $\mathcal{E}_k^t = CATE(x), \; x \sim \mathcal{D}_k^t$ is modeled a client-specific Gaussian random variable
\begin{equation*}
    \mathcal{E}_k^t \sim \mathcal{N}(\mu_k^t, \Sigma_k^t)
\end{equation*}

In adaptation to the drifting task embedding space resulting from the training process, we do not keep the running estimates of the Gaussian statistics. Instead, the estimation is handled by the server, as we will discuss later in the server section.

\noindent \textbf{AC-GAN. } Classical GAN models have two components: generator $G$ and discriminator $D$. They are trained adversarially, where generator $G$ synthesizes fake images and discriminator $D$ distinguishes whether an image is Real or Fake. This means $D$ typically only has a binary R/F head. AC-GAN stands out by adding an auxiliary classification head to the discriminator $D$ to predict the class label of given images. This architecture also enables the generator $G$ to synthesize images exclusive to any given class label $y$. 

\noindent \textbf{Client Training. } Originally, AC-GAN makes predictions from data features only. We expand the input size of the class head in the discriminator $D$, making it capable of processing data features $\mathcal{F}$ concatenated with task embeddings $\mathcal{E}$. 

Given a batch of real data points $B = \{(x_i,y_i)\}$, we first train CATE and discriminator $D$ using Binary Cross-Entropy loss on the output of Real/Fake head and Cross-Entropy loss on the output of class head. Next, we use the generator $G$ to synthesize $|B|$ number of images with class labels $y_i$ from the input batch. After that, we again use Binary Cross-Entropy and Cross-Entropy losses to update the parameters of both AC-GAN and CATE. By the end of each training step, client models rehearse previous knowledge using synthesized data with class label $y$ sampled from all the classes seen at that time. 

\subsection{FedGTEA: Server}
The server has two steps. First, the initial model aggregation yields the initial global model. Then, the regularization and consolidation step optimizes the initial model based on the server loss. 

\noindent \textbf{Initial Model Aggregation. }
After collecting client models $\theta_k^t$, the server first aggregates an initial global model $\hat{\theta}_g^t$ following the principles of FedAvg:
\begin{equation}
    \hat{\theta}_g^t = \sum_{k=1}^N w_k\theta_k^t
\end{equation}
where weights $w_k$ are proportional to the number of local data points $|\mathcal{D}_k^t|$.

However, due to statistical heterogeneity across clients, a naively aggregated global model like $\hat{\theta}_g^t$ is often weak \citep{kairouzAdvancesOpenProblems2021, birashkFederatedContinualLearning2025}. To address this problem, we use an additional regularization step to mitigate forgetting and consolidate the model.

\noindent \textbf{Model Regularization and Consolidation. }
We propose a model regularization and consolidation step on the server to regularize $\hat{\theta}_g^t$ at both data and task levels. Through this step, we want to (i) transfer old knowledge from the previous global model, (ii) promote inter-task separation, (iii) while staying close enough to the initial aggregation $\hat{\theta}_g^t$.

\emph{Knowledge Transfer} can be achieved with the knowledge distillation loss. In order to migrate old knowledge from the previous global model, we use the KL-divergence to match the output probability between the current model $\theta$ and the previous global model $\theta_g^{t-1}$. As a result, suppose the current task identity is $\mathcal{T}^T$, then knowledge distillation loss is formulated as:
\begin{equation}
    \mathcal{L}_{KD} = \sum_{x, y \in A_T} KL(\theta_g^{T-1}(x) \| \theta(x))
\end{equation}
where $A_T$ is a dataset synthesized at the server. Its construction will be mentioned by the end of this subsection.

\emph{Task Separation} needs a way to measure the distance between tasks. Using the synthesized dataset $A_T$, we first split it into non-overlapping subsets $\mathcal{A}_T^t$, where $\mathcal{A}_T^t$ contains data points exclusive to task $\mathcal{T}^t$. Now, we can estimate the mean vector and the covariance matrix of tasks $t=1,2,\ldots,T$:
\begin{equation*}
    \mu_T^t = \text{Avg}(CATE(x_i)),\;\; \Sigma_T^t = \text{Cov}(CATE(x_i))
\end{equation*}
where $x_i$ are data points in $\mathcal{A}_T^t$.
This means, at the current step $T$, the Gaussian distribution of task $t$ is $\mathcal{N}(\mu_T^t, \Sigma_T^t)$. Next, all we need is a way to measure the distance between these Gaussians. 

We choose the 2-Wasserstein to measure the distance between task embeddings. Specifically, given two Gaussians $m_i \sim \mathcal{N}(\mu_i, \Sigma_i)$ $i=1,2$, the 2-Wasserstein distance has a closed form, a more detailed discussion will be included in the appendix.
\begin{align*}
    W_2^2(m_1, m_2) = 
    &\;\|\mu_1-\mu_2\|_2^2  \;+\\ 
&\operatorname{tr}\!\Big(\Sigma_1+\Sigma_2
- 2\,\big(\Sigma_2^{1/2}\,\Sigma_1\,\Sigma_2^{1/2}\big)^{1/2}\Big)
\end{align*} 
Using the above components, we formulate the Wasserstein loss function:
\begin{equation}
    \mathcal{L}_{Wasserstein} = \big[\sum_{1\leq i<j \leq T} W_2^2(\mathcal{N}_i, \mathcal{N}_j) \big]^{-1}
\end{equation}
where $\mathcal{N}_i$ and $\mathcal{N}_j$ are the Gaussians of task $1 \leq i < j \leq T$.

\emph{Drift Retention }means we do not want the new global model to drift far away from the initial aggregation $\hat{\theta}_g^t$. We use the $L_2$ norm to measure the distance between the current model and the initial aggregation. Therefore, we can write the anchor loss in the following:
\begin{equation}
    \mathcal{L}_{anchor} = \| \theta - \hat{\theta}_g^t \|_2
\end{equation}

\emph{Server Loss }$\mathcal{L}_{server}$ is the linear combination of the three components above:
\begin{equation}
    \mathcal{L}_{server} = \alpha \mathcal{L}_{KD} + \beta \mathcal{L}_{Wasserstein} + \gamma \mathcal{L}_{anchor}
\end{equation}
Here $\alpha, \beta, \gamma \in \mathbb{R}$ are all tunable hyperparameters, the detailed discussion is included in the appendix. The global model is obtained by solving the following optimization problem using gradient descent:
\begin{equation}
    \theta_g^t = \arg\min_{\theta \in \Theta} \mathcal{L}
\end{equation}
We will discuss

Finally, the new global model $\theta_g^t$ is distributed to all clients as the starting point of the next training round. 

\noindent \textbf{Synthesized Dataset $\mathbf{A_T}$. } Given a dataset size budget $n^T_k$ for client $\mathcal{C}_k$, generator $G_k^T$ generates synthesized dataset $A_{T}(k)=\{(x_i,y_i)\}$ with $n_k^T$ data points. $y_i$ are sampled uniformly from the seen class labels. Together, their union forms the synthesized dataset
\begin{equation}
    A_T = \bigsqcup_{k=1}^N A_{T}(k)
\end{equation}
Client-side budgets $n_k^T$ are proportional to the size of the local training data size $|\mathcal{D}_k^T|$.

\subsection{Discussion and Remarks}

A \emph{cardinality-agnostic} encoder accepts any number of examples—one, a few, or many—and outputs a coherent task embedding without changing the architecture or retraining. Statistically, tasks differ by their data distributions $p_t(x)$. More samples give a better estimate of $p_t(x)$. As the input set grows, the embedding improves; with only a few points, it still provides a usable, though noisier, representation.

In practice, clients hold very different numbers of examples due to behavior, privacy, and availability. Fixing the input size wastes data when it is abundant and breaks when it is scarce. A cardinality-agnostic encoder is robust across clients, improves accuracy as cardinality increases, and avoids per-client tuning. It follows the simple principle that more data enables better inference.

We compare Gaussian task embeddings with the 2-Wasserstein distance because (i) it has a closed form for Gaussians; (ii) it measures both mean differences $(\mu)$ and the Bures alignment cost between covariances $(\Sigma)$ \citep{bhatia2019bures}; and (iii) it is a true metric.

\noindent \textbf{Contribution remark.} We use AC-GAN as the backbone because it compactly combines replay and classification. Although AC-GAN has been used in FCIL, our contributions are new: we infer task knowledge with CATE and integrate Gaussian task embeddings with the Wasserstein distance.

\section{Experiments}



\begin{table*}[t]
  \centering
  \label{tab:avg-acc-forget}
  \begin{adjustbox}{max width=\textwidth}
  \begin{tabular}{l *{6}{S[table-format=2.1(1.1)]}}
    \toprule
    \multicolumn{1}{c}{\textbf{Model}} &
    \multicolumn{2}{c}{\textbf{Sequence 1: CIFAR10}} &
    \multicolumn{2}{c}{\textbf{Sequence 2: CIFAR100}}&
    \multicolumn{2}{c}{\textbf{Sequence 3: CIFAR100 Superclass}}\\
    \cmidrule(lr){2-3}\cmidrule(lr){4-5}\cmidrule(lr){6-7}
     & {\textbf{Accuracy}$\uparrow$} & {\textbf{Forgetting}$\downarrow$}
     & {\textbf{Accuracy}$\uparrow$} & {\textbf{Forgetting}$\downarrow$}
     & {\textbf{Accuracy}$\uparrow$} & {\textbf{Forgetting}$\downarrow$} \\
    \midrule
    FedAvg               & \num{26.2 +- 2.6} & \num{8.5 +- 1.7} & \num{23.4 +- 2.9} & \num{9.2 +- 1.9} &
      \num{23.7 +- 2.5} & \num{13.2 +- 1.6} \\
    FedProx              & \num{26.1 +- 1.8} & \num{8.6 +- 1.3} & \num{24.1 +- 1.9} & \num{8.4 +- 2.0} &
      \num{23.1 +- 1.9} & \num{14.5 +- 2.3} \\
    \midrule
    ACGAN+FedAvg  & \num{30.5 +- 1.8} & \num{6.9 +- 1.3} & \num{28.1 +- 1.3} & \num{7.7 +- 1.4} &
      \num{27.9 +- 1.3} & \num{10.2 +- 0.9} \\
    ACGAN+FedProx & \num{31.1 +- 1.6} & \num{6.0 +- 1.4} & \num{27.5 +- 0.3} & \num{6.4 +- 1.0} &
      \num{28.9 +- 1.8} & \num{11.1 +- 1.6} \\
    \midrule
    FLwF2T               & \num{29.6 +- 0.9} & \num{7.7 +- 1.1} & \num{30.2 +- 0.7} & \num{7.2 +- 1.8} &
      \num{29.9 +- 1.0} & \num{9.2 +- 1.3} \\
    FedCIL               & \num{32.4 +- 1.9} & \num{6.9 +- 1.9} & \num{31.5 +- 0.4} & \num{7.4 +- 1.2} &
      \num{31.2 +- 1.6} & \num{10.8 +- 2.0} \\
    GLFC                 & \num{35.7 +- 1.1} & \num{6.3 +- 0.9} & \num{33.1 +- 0.6} & \num{10.7 +- 1.8} &
      \num{33.6 +- 1.7} & \num{11.2 +- 2.2} \\
    \addlinespace[2pt]
    \textbf{FedGTEA}     &
      {\bfseries \num{37.1 +- 0.7}} &
      {\bfseries \num{4.5 +- 0.5}}  &
      {\bfseries \num{35.9 +- 0.6}} &
      {\bfseries \num{6.6 +- 1.7}}  &
      {\bfseries \num{35.1 +- 1.2 }} &
      {\bfseries \num{8.6 +- 1.4}} \\
    \bottomrule
  \end{tabular}
  \end{adjustbox}
  \caption{Average test accuracy (\(\uparrow\)) and average forgetting (\(\downarrow\)) under federated class-incremental learning on Sequence 1:CIFAR10, Sequence 2:CIFAR100, and Sequence 3:CIFAR100 Superclass. Each dataset is split into a sequence of disjoint class-incremental tasks; clients observe local streams and periodically synchronize with a central server. Forgetting is computed as the per-class drop between the best and final accuracy, averaged over classes. Results are reported as mean\(\pm\)sd over multiple random seeds.}
\end{table*}
In this section, we compare our algorithm FedGTEA with other powerful baselines over average accuracy and average forgetting. We first explain our experiment settings, including datasets, task sequences, and federation settings. 

\subsection{Baselines}
We evaluate FedGTEA against representative methods across FL, CIL, and FCIL. For FL baselines, we include \textbf{FedAvg}\citep{mcmahanCommunicationEfficientLearningDeep2017}, a standard averaging-based method, and \textbf{FedProx}\citep{liFederatedOptimizationHeterogeneous2020}, which introduces a proximal term to handle client heterogeneity. For CIL, we consider \textbf{iCaRL} \citep{rebuffiICaRLIncrementalClassifier2017}, which maintains exemplar memory and uses prototype-based classification, and \textbf{DER} \citep{buzzegaDarkExperienceGeneral2020}, which mitigates forgetting through logit alignment via knowledge distillation. Within FCIL, we compare against \textbf{FLwF2T} \citep{usmanovaDistillationbasedApproachIntegrating2021}, which transfers knowledge across clients and rounds via distillation; \textbf{FedCIL} \citep{qiBetterGenerativeReplay2023}, which adapts ACGAN-based generative replay to the federated setting; and \textbf{GLFC} \citep{dongFederatedClassIncrementalLearning2022a}, which combines local exemplar rehearsal with global knowledge alignment to reduce forgetting. These baselines collectively span a wide range of strategies, including averaging/proximal optimization, exemplar–prototype rehearsal, logit-based distillation, and generative replay, providing a strong and comprehensive benchmark for FCIL evaluation.

\subsection{Experimental Settings \& Evaluation}

\noindent \textbf{Datasets.}
We evaluate FedGTEA and baselines on three standard FCIL datasets \citep{krizhevsky2009learning} \textsc{CIFAR-10}, \textsc{CIFAR-100} icarl split \citep{rebuffiICaRLIncrementalClassifier2017}, and \textsc{CIFAR-100} superclass split. 

\noindent \textbf{CIFAR100 Superclass. } The 100 classes of \textsc{CIFAR100} are grouped into 20 non-overlapping \emph{superclasses}, each containing five semantically related classes. This offers a natural and semantically meaningful task split option.


\noindent \textbf{Federation \& Task Sequences.}
We have three task sequences. One is for CIFAR10, with two different task sequences for CIFAR100. For CIFAR10, we follow the settings set by \citep{qiBetterGenerativeReplay2023}. For CIFAR100, the two sequences follow the principles set by \citep{rebuffiICaRLIncrementalClassifier2017} and \citep{yoonFederatedContinualLearning2021}.
\begin{itemize}
  \item \textbf{Sequence 1:} CIFAR10. There are 5 clients with 5 tasks. Each task contains 2 non-overlapping classes. Classes are randomly selected. 
  \item \textbf{Sequence 2:} CIFAR100. Using the icarl task split, we use 10 clients with 10 tasks. The task split was randomly configured by setting a fixed random seed. Each task has 10 non-overlapping classes.
  \item \textbf{Sequence 3:} CIFAR100. With the Superclass task split, we configure 10 clients and take each superclass as a task. Therefore, we have in total 20 tasks. Each task has 5 semantically related classes. 
\end{itemize}

\noindent \textbf{Metrics.}
We report \emph{average accuracy} (higher is better) and \emph{average forgetting} (lower is better).
Average forgetting measures how much performance on a past task drops after learning later tasks: for each task we compute the gap between its best (peak) accuracy and its accuracy at the end of training, and then average across tasks.

\subsection{Results \& Analysis}
FedGTEA advances the accuracy–forgetting trade-off on all benchmarks, yielding higher final accuracy while keeping interference low.
\begin{itemize}
  \item \emph{Results on Sequence 1:} FedGTEA achieves the highest accuracy (37.1\(\pm\)0.7) and the only sub-5\% forgetting (4.5\(\pm\)0.5); other methods remain at or above 6\% forgetting.
  \item \emph{Results on Sequence 2:} FedGTEA attains the best accuracy (35.9\(\pm\)0.6) while maintaining single-digit forgetting (6.6\(\pm\)1.7), competitive with the lowest-forgetting baseline (6.4) but at substantially higher accuracy.
  \item \emph{Results on Sequence 3:} FedGTEA delivers both the best accuracy (35.1\(\pm\)1.2) and the lowest forgetting (8.6\(\pm\)1.4); baselines typically lie in the 9–14\% forgetting range.
\end{itemize}

Incorporating task context during aggregation aligns client updates and curbs cross-task interference beyond what rehearsal- or distillation-only baselines achieve, leading to consistent gains across datasets.








\section{Conclusion}


In this study, we propose FedGTEA, an algorithm that models task-level information in a scalable and efficient manner. On the client side, we introduce a parameter-efficient Cardinality-Agnostic Task Encoder (CATE) to infer task embeddings, modeling them as Gaussian random variables to capture task position and uncertainty. On the server side, we apply the 2-Wasserstein distance to regularize task representations and promote inter-task separation. Empirical results demonstrate that our proposed method outperforms strong baselines, achieving higher average accuracy and reduced forgetting.

\section*{ACKNOWLEDGMENT}

This research was supported by the Office of Naval Research (ONR) grant N000142312629.

\bibliographystyle{plainnat}
\bibliography{references}

\clearpage
\appendix
\onecolumn

\section*{Appendix}

This appendix provides supplementary material to support the main paper. It complements the theoretical discussion and experimental results of this paper. 
\textbf{In section A}, we begin with a detailed theoretical discussion of the 2-Wasserstein distance, covering its computation and justifying its selection over alternative metrics. 
\textbf{In section B}, we then present a comprehensive overview of our experimental setup, including descriptions of the datasets, baseline models, performance metrics, and hyperparameter configurations. 
\textbf{In section C}, we present an extensive ablation study that systematically evaluates the contribution of each key component of our proposed FedGTEA framework. This study demonstrates that removing any part of the model—specifically the CATE task encoder, the Wasserstein loss, the anchor loss, or the distillation loss—results in a degradation of performance. These findings validate our design choices and underscore the synergistic effect of the components in achieving the reported results.

\section{Wasserstein Distance}

\subsection{Computation and Complexity}
Wasserstein distance is a core component of our model consolidation step. Given any two tasks $1 \leq i < j \leq T$, their task embeddings are distributed as Gaussian distributions. We denote these Gaussian distributions as $\mathcal{N}_i = \mathcal{N}(\mu_i, \Sigma_i)$ and $\mathcal{N}_j = \mathcal{N}(\mu_j, \Sigma_j)$. The distance between these two tasks, which is proxied by the 2-Wasserstein distance between the corresponding two Gaussian distributions, can be computed in the following closed form \citep{peyreComputationalOptimalTransport2020}: 
\begin{equation*}
    W_2^2(\mathcal{N}_i, \mathcal{N}_j) = \| \mu_i - \mu_j \|_2^2 + \text{tr}(\Sigma_i + \Sigma_j - 2(\Sigma_i^{1/2} \Sigma_j \Sigma_i^{1/2})^{1/2})
\end{equation*}
where $\| \cdot \|_2$ is the Euclidean norm and $\Sigma^{1/2}$ is the matrix square root such that $(\Sigma^{1/2})^2 = \Sigma^{1/2}\Sigma^{1/2} = \Sigma$. It is important to note that in our case, $\Sigma$ as the covariance matrix is Positive Semi-Definite (PSD). This means we can use eigen-decomposition for symmetric matrices to compute the square root in an efficient and numerically stable manner. Previous work has proved a computational complexity $\mathcal{O}(n^3)$ \citep{peyreComputationalOptimalTransport2020} for calculating the 2-Wasserstein distance between two Gaussian distributions. 

\subsection{Why Wasserstein?}

In this part, we compare the Wasserstein distance with the other two popular choices. We argue that, as a genuine mathematical metric without an upper bound on the distance, the 2-Wasserstein distance is the overall best option for our case. 

While several metrics exist for comparing Gaussian distributions, $\mathcal{N}_i(\mu_i, \Sigma_i)$ and $\mathcal{N}_j(\mu_j, \Sigma_j)$, two common alternatives to the Wasserstein distance are the Kullback-Leibler (KL) Divergence and the Bhattacharyya Distance. Their closed-form expressions are given by:
$$D_{KL}(\mathcal{N}_i || \mathcal{N}_j) = \frac{1}{2} \left( \text{tr}(\Sigma_j^{-1}\Sigma_i) + (\mu_j - \mu_i)^T \Sigma_j^{-1} (\mu_j - \mu_i) - k + \ln\left(\frac{\det(\Sigma_j)}{\det(\Sigma_i)}\right) \right)$$
$$D_B(\mathcal{N}_i, \mathcal{N}_j) = \frac{1}{8}(\mu_i - \mu_j)^T \Sigma_{ij}^{-1} (\mu_i - \mu_j) + \frac{1}{2} \ln\left(\frac{\det(\Sigma_{ij})}{\sqrt{\det(\Sigma_i)\det(\Sigma_j)}}\right)$$
where $\Sigma_{ij} = \frac{\Sigma_i + \Sigma_j}{2}$. All three distances (Wasserstein, KL, Bhattacharyya) share a similar computational complexity of $\mathcal{O}(n^3)$ due to matrix operations like determinant calculation. 
The choice of Wasserstein distance is motivated by the following drawbacks of the alternatives.

KL-Divergence is not a true mathematical metric because it is asymmetric, meaning $D_{KL}(\mathcal{N}_i || \mathcal{N}_j) \neq D_{KL}(\mathcal{N}_j || \mathcal{N}_i)$. This property is undesirable for our objective function, as it introduces a sensitivity to the sequence of tasks. Model performance should be invariant to task order, a requirement that the asymmetry of the KL-Divergence violates. Moreover, although Bhattacharyya Distance is a true metric, it is bounded and normalized. While useful as a similarity score, an upper bound is problematic for a loss function. Unlike unbounded losses such as Cross Entropy and MSE, which can generate large corrective gradients when a model is far from the optimum. The bounded nature of the Bhattacharyya distance can lead to weak gradients and slow convergence in such scenarios. 

\noindent \textbf{Wasserstein Distance} provides a powerful alternative that resolves the issues mentioned above. It is intuitively understood as the minimum cost to transform one distribution into another. First, the Wasserstein distance provides a meaningful and smooth measure even for distributions that are far apart or have non-overlapping support. For KL-Divergence, the value can become infinite if the distributions do not overlap, leading to vanishing or exploding gradients. The Wasserstein distance, however, always provides a finite value and a usable gradient, resulting in a smoother and more stable loss landscape. 

Second, for Gaussian distributions specifically, the 2-Wasserstein distance has a particularly elegant geometric interpretation. The formula separates the distance into a Euclidean distance between the means and a trace norm involving the covariance matrices. This means it cleanly measures the difference in the location and the shape of the distributions, reflecting the geometry of the underlying parameter space.

\section{Experiments. }

\subsection{Datasets}

\noindent \textbf{CIFAR10} \citep{krizhevsky2009learning} is an image classification dataset of 10 classes with 60,000 instances. The training split has 50,000 images, and the test split contains the rest 10,000. Different classes have the same number of images in the dataset. In other words, all classes have 500 data points in the train split. The situation is the same with the test split, with 200 images per class. 

\noindent \textbf{CIFAR100} \citep{krizhevsky2009learning} is also an image classification dataset. Similar to CIFAR10, each class has 500 training data points with 100 test cases. One significant difference is that CIFAR100 comes with a coarse Superclass label. Each superclass comprises 5 distinct classes, and different superclasses do not overlap. A detailed superclass split can be found on the CIFAR100 website, as shown in the table below. Task split based on superclass is more semantically meaningful than random split based iCaRL split, which splits the tasks by setting a random seed 1993. Therefore, the distinction between tasks is clearer for the superclass split. Moreover, it supplements an additional federation configuration with 20 tasks and 5 classes per task to the typical 10 tasks with 10 classes per task. 

\begin{center}
    
\begin{tabular}{ll}
\toprule
\textbf{Superclass} & \textbf{Classes} \\
\midrule
aquatic mammals & beaver, dolphin, otter, seal, whale \\
fish & aquarium fish, flatfish, ray, shark, trout \\
flowers & orchids, poppies, roses, sunflowers, tulips \\
food containers & bottles, bowls, cans, cups, plates \\
fruit and vegetables & apples, mushrooms, oranges, pears, sweet peppers \\
household electrical devices & clock, computer keyboard, lamp, telephone, television \\
household furniture & bed, chair, couch, table, wardrobe \\
insects & bee, beetle, butterfly, caterpillar, cockroach \\
large carnivores & bear, leopard, lion, tiger, wolf \\
large man-made outdoor things & bridge, castle, house, road, skyscraper \\
large natural outdoor scenes & cloud, forest, mountain, plain, sea \\
large omnivores and herbivores & camel, cattle, chimpanzee, elephant, kangaroo \\
medium-sized mammals & fox, porcupine, possum, raccoon, skunk \\
non-insect invertebrates & crab, lobster, snail, spider, worm \\
people & baby, boy, girl, man, woman \\
reptiles & crocodile, dinosaur, lizard, snake, turtle \\
small mammals & hamster, mouse, rabbit, shrew, squirrel \\
trees & maple, oak, palm, pine, willow \\
vehicles 1 & bicycle, bus, motorcycle, pickup truck, train \\
vehicles 2 & lawn-mower, rocket, streetcar, tank, tractor \\
\bottomrule
\end{tabular}
\end{center}

\subsection{Baselines}

We compare our method FedGTEA, with two baselines from FL, one from CL, and three from FCIL. The plain FL methods simply train a global model on a sequence of tasks, without any modifications like memory. The CL method uses the AC-GAN module to replay and classify, which leverages generative replay to fight catastrophic forgetting. The three popular baselines in FCIL focus on addressing both catastrophic forgetting and statistical heterogeneity across clients. 

\noindent \textbf{FedAvg }\citep{mcmahanCommunicationEfficientLearningDeep2017}. As a representative FL algorithm, FedAvg trains client models with local datasets and aggregates client models in a weighted sum manner. The weights we proportional to the number of data points in the local datasets. 

\noindent \textbf{FedProx }\citep{liFederatedOptimizationHeterogeneous2020}. In addition to FedAvg's aggregation approach, FedProx adds a regularization term in the client's local training process. To avoid significant divergence in update directions across client models, clients are penalized for deviating from the last round's global model. This regularization controls the degree of deviation from the previous global round. 

\noindent \textbf{AC-GAN-Replay }\citep{wu2018memory}. This algorithm employs a GAN-based generative replay method. In addition to a traditional GAN's Real or Fake binary classification head, AC-GAN has an auxiliary classification head for classes. This enables its generator to synthesize images exclusive to any selected class. 

\noindent \textbf{FLwF2T }\citep{usmanovaDistillationbasedApproachIntegrating2021}. FLwF2T is an FCIL algorithm that adopted knowledge distillation within the FL framework. It transfers knowledge from both the previous classifier from the previous task and the last round's global classifier to the current one. 

\noindent \textbf{FedCIL }\citep{qiBetterGenerativeReplay2023}. This FCIL algorithm extends the AC-GAN-assisted FL framework one step further by adding an additional feature alignment and model consolidation step on the server. With distillation techniques embedded, it delivers more robust results. 

\noindent \textbf{GLFC }\citep{dongFederatedClassIncrementalLearning2022a}. Under the FCIL scenario, GLFC utilizes a distillation-based method together with a memory buffer to store previous representative data points. Although GLFC does not fall under the strict FCIL because it uses raw data from previous tasks, it alleviates catastrophic forgetting from both local and global perspectives. 

\subsection{Performance Metrics}

We use the metrics of average accuracy and average forgetting to evaluate the performance of our model and baselines \citep{yoonFederatedContinualLearning2021, mirzadeh2020linear}. Suppose $a_k^{t,i}$ is the test accuracy of the $i$-th task after learning the $t$-th task in client $k$.

\noindent \textbf{Average Accuracy. }The final metric is computed after the training phase. We calculate the test accuracy for all seen tasks for all clients. The weighted sum uses the number of data points in the local dataset as weights:

$$
\text{Average Accuracy} = \frac{1}{\sum_{k=1}^{N} \sum_{i=1}^{T} n_{k}^{i}} \sum_{k=1}^{N} \sum_{i=1}^{T} a_{k}^{T,i} * n_{k}^{i}.
$$

Here $n_k^i$ is the number of data points of client $k$'s train dataset at task $i$. This enables a fair evaluation accounting for the variation in task difficulty across clients. 

\noindent \textbf{Average Forgetting. }This metric assesses the degree of backward transfer during the continual learning phase. By design, it calculates the difference between the peak accuracy and the ending accuracy of each task for each client. Weighted sum is also used in the formulation:
$$
\text{Average Forgetting} = \frac{1}{\sum_{k=1}^{N} \sum_{i=1}^{T-1} n_{k}^{i}} \sum_{k=1}^{N} \sum_{i=1}^{T-1} \max_{t \in \{1,...,T-1\}} (a_{k}^{t,i} - a_{k}^{T,i}) * n_{k}^{i}.
$$

\subsection{Hyperparameter Configuration}

\noindent \textbf{Optimization Details.}
We employed the Adam (Adaptive Moment Estimation) gradient descent optimizer for all training procedures. It combines the advantages of two other popular methods: the momentum technique, which accelerates convergence, and RMSprop, which adapts the learning rate based on the magnitude of recent gradients. A consistent batch size of 64 was used across all experiments. For the CIFAR-10 dataset, we set the learning rate to $1 \times 10^{-4}$ and trained for 60 global communication rounds, with each client performing 100 local iterations per round. For the more complex CIFAR-100 task splits, the learning rate was increased to $1 \times 10^{-3}$, with training conducted over 40 global rounds and 400 local iterations to account for the larger dataset. During client-side training, the number of synthesized images was set to match the batch size (64). In the server-side regularization step, each generator was allocated a budget to produce 200 data points for each class observed in previous tasks.

\noindent \textbf{Model Architectures.}
For fair comparison, all models utilize a common Convolutional Neural Network (CNN) architecture as the feature extractor, trained from scratch. This CNN consists of six convolutional layers with channel sizes of $[16, 32, 64, 128, 256, 512]$. The generator model takes a 100-dimensional random noise vector, concatenated with a one-hot class vector, as input. This input is first projected by a fully-connected layer into a 384-dimensional vector, which is then passed through four transposed convolutional layers with channel sizes of $[384, 192, 96, 48, 3]$ to produce an image. The discriminator is composed of two fully-connected heads: one performing binary classification (real vs. fake) and another performing multi-class classification to identify the image's class.

\noindent \textbf{Model Regularization.}
The server-side regularization is governed by a composite loss function ($\mathcal{L}_{\text{server}}$) which combines three distinct terms: an anchor loss ($\mathcal{L}_{\text{anchor}}$) to penalize drastic model updates, a Wasserstein loss ($\mathcal{L}_{\text{Wasserstein}}$) to enhance inter-task feature separation, and a knowledge distillation loss ($\mathcal{L}_{\text{KD}}$) to transfer knowledge from previous models. The total loss is formulated as:
\begin{equation*}
    \mathcal{L}_{\text{server}} = \alpha \mathcal{L}_{\text{KD}} + \beta \mathcal{L}_{\text{Wasserstein}} + \gamma \mathcal{L}_{\text{anchor}}
\end{equation*}
The coefficients $\alpha, \beta,$ and $\gamma$ balance the contribution of each term. While specific configurations could optimize for a single metric (e.g., a large $\alpha$ reduces forgetting), we aimed for a balanced performance. Following a grid search, we selected the configuration $\alpha = 0.3, \beta=0.3,$ and $\gamma=0.4$, which we found provides a favorable trade-off between classification accuracy and catastrophic forgetting.

\section{Ablation Study}

Our proposed method FedGTEA has two major components: the client model consists of an AC-GAN module assisted by CATE from a task perspective, and the server side leverages the model consolidation and regularization step to enforce more robust performances with low variances. The two major novelties of our paper are the task encoder CATE and the model consolidation and regularization step. More precisely, the regularization comprises three loss functions: anchor loss, distillation loss, and Wasserstein loss. We conduct ablation studies over all three tested scenarios by presenting model performance without certain parts. As shown in the following table, losing any parts of FedGTEA hurts the efficacy of our model. Specifically, we notice a big decrease in performance over CIFAR100 superclass split without the Wasserstein loss and CATE task embedding module, which further proves our claim that CATE + regularization provides robust task knowledge.

\begin{itemize}
  \item \textbf{Overall effectiveness.} The ablation study in Table~1 shows that the full FedGTEA consistently attains the highest accuracy and the lowest forgetting across all three experimental sequences, outperforming every ablated variant. This establishes the complete framework as the strongest configuration and sets the reference point for evaluating the impact of removing individual components.

  \item \textbf{Distillation and anchor losses.} Removing specific losses markedly harms performance, revealing their distinct roles. Eliminating the distillation loss induces the largest rise in catastrophic forgetting, with the forgetting metric on \textsc{CIFAR100} Superclass increasing by approximately 42\% (from 8.6 to 12.2), underscoring its importance for retaining prior knowledge. Dropping the anchor loss leads to a pronounced accuracy decline, including a nearly 7\% absolute drop on \textsc{CIFAR10}, indicating its necessity for stable and discriminative feature representations.

  \item \textbf{CATE and Wasserstein effects.} The absence of the CATE module and the Wasserstein loss also yields considerable degradation, with accuracies in some cases falling to levels comparable to or even below the GLFC baseline. Taken together, these results validate the design choices: the synergy between the CATE task encoder and the combined regularization losses—anchor, distillation, and Wasserstein—is essential for achieving the state-of-the-art performance of FedGTEA.
\end{itemize}


\begin{table*}[h]
  \centering
  \label{tab:avg-acc-forget}
  \caption{Ablation study of FedGTEA components. We report the average accuracy (\%) and forgetting (\%) over 5 runs. The best results are in \textbf{bold}. For accuracy, higher is better ($\uparrow$). For forgetting, lower is better ($\downarrow$).}
  \begin{adjustbox}{max width=\textwidth}
  \begin{tabular}{l *{6}{S[table-format=2.1(1.1)]}}
    \toprule
    \multicolumn{1}{c}{\textbf{Model}} &
    \multicolumn{2}{c}{\textbf{Sequence 1: CIFAR10}} &
    \multicolumn{2}{c}{\textbf{Sequence 2: CIFAR100}}&
    \multicolumn{2}{c}{\textbf{Sequence 3: CIFAR100 Superclass}}\\
    \cmidrule(lr){2-3}\cmidrule(lr){4-5}\cmidrule(lr){6-7}
     & {\textbf{Accuracy}$\uparrow$} & {\textbf{Forgetting}$\downarrow$}
     & {\textbf{Accuracy}$\uparrow$} & {\textbf{Forgetting}$\downarrow$}
     & {\textbf{Accuracy}$\uparrow$} & {\textbf{Forgetting}$\downarrow$} \\
    \midrule
    FLwF2T               & \num{29.6 +- 0.9} & \num{7.7 +- 1.1} & \num{30.2 +- 0.7} & \num{7.2 +- 1.8} &
      \num{29.9 +- 1.0} & \num{9.2 +- 1.3} \\
    FedCIL               & \num{32.4 +- 1.9} & \num{6.9 +- 1.9} & \num{31.5 +- 0.4} & \num{7.4 +- 1.2} &
      \num{31.2 +- 1.6} & \num{10.8 +- 2.0} \\
    GLFC                 & \num{35.7 +- 1.1} & \num{6.3 +- 0.9} & \num{33.1 +- 0.6} & \num{10.7 +- 1.8} &
      \num{33.6 +- 1.7} & \num{11.2 +- 2.2} \\
    \midrule
    \addlinespace[2pt]
    \textbf{FedGTEA w/o CATE \& Wasserstein} & \num{32.6 +- 0.5} & \num{7.1 +- 0.7} & \num{32.2 +- 0.5} & \num{8.1 +- 1.1} & \num{31.7 +- 0.7} & \num{10.5 +- 0.9} \\
    \textbf{FedGTEA w/o Wasserstein} & \num{34.1 +- 0.7} & \num{5.8 +- 0.4} & \num{33.3 +- 0.4} & \num{8.8 +- 0.7} & \num{32.2 +- 0.3} & \num{10.3 +- 0.3}   \\
    \textbf{FedGTEA w/o Anchor} & \num{30.2 +- 1.3} & \num{6.9 +- 1.4} & \num{32.5 +- 0.4} & \num{8.1 +- 0.3} & \num{31.0 +- 0.4} & \num{10.8 +- 0.2} \\
    \textbf{FedGTEA w/o Distillation} & \num{32.3 +- 1.5} & \num{8.7 +- 1.1} &  \num{31.9 +- 0.6} & \num{10.9 +- 1.6} & \num{31.4 +- 1.1} & \num{12.2 +- 2.4}\\
    \textbf{FedGTEA}     &
      {\bfseries \num{37.1 +- 0.7}} &
      {\bfseries \num{4.5 +- 0.5}}  &
      {\bfseries \num{35.9 +- 0.6}} &
      {\bfseries \num{6.6 +- 1.7}}  &
      {\bfseries \num{35.1 +- 1.2 }} &
      {\bfseries \num{8.6 +- 1.4}} \\
    \bottomrule
  \end{tabular}
  \end{adjustbox}
\end{table*}














\end{document}